# Benchmarking Vision Language Models on German Factual Data

René Peinl and Vincent Tischler

`rene.peinl@hof-university.de` - `vincent.tischler.3@hof-university.de`
Hof University of Applied Sciences
Germany

**Abstract:** Similar to LLMs, the development of vision language models is mainly driven by English datasets and models trained in English and Chinese language, whereas support for other languages, even those considered high-resource languages such as German, remains significantly weaker. In this work we present an analysis of open-weight VLMs on factual knowledge in the German and English language. We disentangle the image-related aspects from the textual ones by analyzing accuracy with jury-as-a-judge in both prompt languages and images from German and international contexts. We found that for celebrities and sights, VLMs struggle because they are lacking visual cognition of German image contents. For animals and plants, the tested models can often correctly identify the image contents according to the scientific name or English common name but fail in German language. Cars and supermarket products were identified equally well in English and German images across both prompt languages.

## 1 Introduction / Motivation

Building on the success of large language models (LLMs), a new category of deep learning architectures was proposed – vision language models (VLMs). These models combine pre-trained image encoders with pre-trained LLMs, making them well-suited for multi-modal tasks like visual question answering (VQA). The amount of research papers and publicly available models published surged [1], [2] since the concept was proven effective with Flamingo [3] and the BLIP family of models [4], [5], [6] as role models. In 2024, open weight models have closed the gap [7] to the big proprietary models like GPT-4o and Claude 3.5 Sonnet, that have been at the top of the leaderboards across a large number of benchmarks [8]. Although large open VLMs with 70 to 90 billion parameters [9], [10], [11] still dominate, even medium-sized VLMs like Molmo [12] and Pixtral [13] with 7 and 12 billion parameters respectively, deliver good performance in popular benchmarks (see Table 2). Similar to LLMs [14], the development of VLMs is mainly driven by English datasets and models trained in English and Chinese language, whereas support for other languages, even those considered high-resource languages such as German, is considerably less pronounced [15], [16]. This led us to the question whether underperformance is mainly due to the language understanding of the underlying LLM or if the finetuning with English-dominated image/text pairs in later stages of the VLM development is equally relevant.

Our hypothesis for the analysis is that it is not enough to use a multi-lingual LLM as a backbone for a VLM to create a multi-lingual VLM. Furthermore, we hypothesize that country-specific contents of images are equally important as language understanding for multi-lingual comprehension.

Our paper is structured as follows. We first analyze related work from multi-lingual and cultural VQA. Then we present the dataset and describe how it was collected. After that, we explain our evaluation methodology before presenting results and discussing our interpretation. We admit some limitations before summarizing our contribution and closing with an outlook.

## 2    Literature Review

Our work lies in the intersection between evaluating generative AI models on multiple languages regarding the differences in output quality and the benchmarking of VLMs. Evaluation datasets for VLMs are abundant [17].

### 2.1    General Benchmarks for Vision Language Models

After first datasets like the VQA [18] (v1 and v2), that can also be used for training, and OK-VQA [19], that requires background knowledge to answer the questions, the evaluations got more and more diverse: Text-VQA [20] concentrates on OCR capabilities of the models. ChartQA [21] evaluates the models abilities to interpret different kinds of charts, which is also relatively OCR-heavy. MathVista [22] includes images as part of mathematical tasks. MMvet [23] focuses on multi-modal reasoning capabilities. MMMU [24] can be seen as the VLM counterpart of the popular MMLU benchmark [25] for LLMs and tests models across a wide range of domains from social sciences to physics. Hallucinations, model answers that sound plausible, but are incorrect, are also an issue for VLMs and can be tested with datasets like HallusionBench [26] or POPE [27]. Comprehensive benchmarks like SEED-Bench [28], [29] and MMBench [30] collect a large number of existing benchmarks and try to balance the tasks so that a wide range of capabilities of the models can be tested in a single benchmark.

### 2.2    Multilingual Benchmarks for VLMs

Most of the above benchmarks are in the English language and only few are available in multiple languages [17], which in turn focus on English and Chinese. **MTVQA** [31] is a notable exception and comprises 6,778 question-answer pairs across 2,116 images for nine languages including German, French and Italian. It seems quite challenging as even the best models only achieved around 30% accuracy in the mentioned languages at the time of publishing the benchmark. The other languages like Japanese, Russian and Thai were even worse. However, they do not compare to English language.

**Exams-V** [32] can be seen as a multi-lingual version of MMMU, covering exams from diverse domains across eleven languages including German, French and Italian. However, the distribution of domains across languages is not homogeneous. For Polish, there is only one domain (business), whereas for Croatian there are 13 domains including physics, biology, history and politics. Only around a fifth of questions are image-related. Astonishingly, English exams can be solved only in 29.3% of cases by the best model GPT-4V whereas Croatian exams were answered correctly in 55.5% and French even in 60.3% of cases. **MMMBench** [33] is another dataset that consists of image text pairs in six languages. Besides English and Chinese, this includes Turkish, Russian, Portuguese and Arabic. Li et al [34] translated a subset of ImageNet, VQA v2 and OK-VQA into 80 languages, including most European languages. **xGQA** [35] is a translation of GQA into seven additional languages, including German. Evaluation results show a gap of 20% and more between English and German language. Most of these datasets rely on machine translation, although Park et al [36] find that translation artefacts can significantly affect the performance of models and can shift the meaning far enough to change the correct answer from yes to no. Therefore, we only use manually verified and quality checked translations, including reverse image search in some cases. All of these datasets focus on text in different languages, not country-specific images.

## 2.3 Intercultural Benchmarks for VLMs

A different, but closely related strand of research deals with cultural aspects of VLM tasks. These refer not only to language-related issues, that are mainly concerned with the textual part and OCR, but also with image-related questions that might be country or culture-specific. Nayak et al. [37] compile **CulturalVQA**, "a visual question-answering benchmark aimed at assessing VLM's geo-diverse cultural understanding". They collect images from the categories traditions, rituals, food, drink, and clothing from eleven countries, including Germany, to ask questions that require not only cultural image understanding, but partly also background knowledge. One question is e.g. showing an image of a Bavarian pretzel and asking about a famous event where this dish is served, which is kind of ambiguous because pretzels are served at most Bavarian folk festivals. Questions are in English only. Romero et al. [15] introduce **CVQA**, a large-scale multilingual VQA benchmark, representing the cultures of 30 countries and languages across 10 categories, with 5,000 images and 10,000 questions. There is a large overlap with our dataset regarding the categories, as they also include animals, plants, products, vehicles, famous people and landmarks. However, they do not include German images or texts. They found that Spanish images are much better recognized than French and Brazilian are better recognized than Japanese. They also found that using English language for prompting and answers leads to better results than using local language. **WorldCuisines** [38] is similar to CVQA, but concentrates on food and underrepresented cultural contexts. It consists of 1.15 million data points from different countries such as Germany, France and Spain. However, they did not include questions in the German language in their dataset. **ALM-Bench** [39] features 12,700 questions from 19 categories in 100 languages from 73 countries. They found that German was together with French, Italian and Spanish among the best recognized languages after

English across all tested VLMs. GLM-4V 9B was the best open weight model followed by Qwen 2 VL. GPT-4o and Gemini 1.5 Pro outperformed the open weight models. They also did a quantitative error analysis categorizing errors into "lack of cultural understanding", "language errors", "lack of knowledge", "reasoning errors", "perceptual errors", and "translation errors".

Our main differentiation is that we concentrate on factual knowledge, do a deeper analysis of errors across categories, use manual translations and compare not only image contents but also prompt and answer language between German and English.

## 3 Dataset

We wanted to test fact-based knowledge instead of asking questions that require deeper understanding or reasoning to focus on the language aspect (German vs. English names) as well as on the country-specific image contents. We ask for perceptual facts like a person's name, the make and model of a car or the species name of an animal.

Table 1. Overview of the dataset used for evaluation[1]

| category | lang | # images | public | remarks and examples |
| --- | --- | --- | --- | --- |
| animals | de | 172 | no | magpie, chamois, brown trout, swallowtail, European adder |
|  | en | 175 | no | greater flamingo, barracuda, gorilla, Komodo dragon, common morpho |
| plants | de | 88 | no | Edelweiss, Alpine gentian, ivy |
|  | en | 97 | yes | oxford-flowers102 (single image per class) |
| celebrities | de | 153 | no | Helene Fischer, Markus Soeder, Toni Kroos, Stefan Raab |
|  | en | 99 | no | Beyoncé, Mike Pence, Usain Bolt, Gal Gadot |
|  | en | 422 | yes | SatyaV/celeb-1000 on huggingface (filtered) |
| sights | de | 45 | no | Cologne Cathedral, Brandenburg Gate |
|  | en | 47 | no | Big Ben, Eiffel tower, Golden Gate bridge |
| cars | de | 164 | no | Audi A4, Skoda Fabia, VW ID3 |
|  | en | 181 | yes | Stanford-cars (filtered for higher resolution) |
| products | de | 88 | no | Rewe whipped cream, Kinder Milch-Schnitte |
|  | en | 106 | no | Dr. Pepper Soda Pop, Ruffles potato chips |
| **Total** | **de** | **710** |  |  |
|  | **en** | **705** |  | Without Celeb1k |

The dataset consists of the categories animals, plants, celebrities, sights, cars and products. Every category consists of a subset with image contents related to Germany and a second subset, that features image contents that stem from the US or are internationally well-known. For animals that would be e.g. animals that are typically seen in

---

[1] Dataset can be downloaded at https://huggingface.co/datasets/rpeinl/FactSightVQA

Zoos worldwide like parrots, lions, koalas, turtles, manta rays and walking leaves. For celebrities, we include US-based people like Taylor Swift or Tom Cruise, but also internationally well-known politicians like Emanuel Macron or athletes like Neymar. Images were taken from public websites, avoiding images from Wikipedia. Image sizes are rather large with a minimum of 400px and a maximum of 1300px (avg 750px). Most parts of the data are self-collected (indicated by public=no in **Table 1**).

Some parts are reused from well-known public datasets like Stanford cars and Oxford flowers. To find out whether there is a difference between the self-collected and public images, we included the dataset Celeb1k in addition to the self-collected international celebrities, which is hosted on Huggingface and therefore with some probability part of the training-data of the tested VLMs. The image size for this dataset is smaller than ours (256px squared). We use a filtered subset with only 211 different people and two images per person. The products category contains product images from online supermarkets that deliver food. The questions related to products can be answered by using OCR and do not need a country-specific image understanding.

## 4 Evaluation Methodology

### 4.1 Choice of VLMs to Test

The focus of our analysis is on medium-sized models with 7B to 13B parameters that are freely available as open weight. These can be easily run on modest hardware. However, since factual questions do not suggest that large LLM backbones will be required to give good answers, we also included a few smaller models (2-4B parameters) as well as a few bigger ones (25B-38B parameters) to compare results across VLM sizes. We concentrated on recently published models that showed strong performance in the popular benchmarks used in the Open VLM leaderboard (OVLM) [8]. We ended up with eight models in the focus area, as well as two small and three big VLMs.

The main evaluation happened in December 2024 and January 2025. We even updated the already processed Qwen 2 VL to its successor Qwen 2.5 VL, which was published 26[th] of January 2025. Aria is an exception regarding two aspects. It is the only model using a Mixture-of-Expert (MoE) architecture [40] instead of a dense one. It further doesn't use a pre-trained LLM but instead uses an approach they call a "native multimodal training", that still consists of text-only training as the first step [41].

As seen in **Table 2**, SigLIP is mainly used as vision encoder, followed by some CLIP ViT variations. Only the biggest model uses an image encoder with more 300-400M parameters. The LLM backbones are relatively diverse, and we purposefully chose some models that allow us to compare different VLM training with the same LLM backbone like PaliGemma2 vs. Ovis1.6 Gemma2 that both use the Gemma2 9B LLM. We also included Ovis1.6 Gemma2 28B to see the influence of a larger LLM on the performance of the VLM when the training data is the same. We always use the instruct version of the VLM if one is available. OVLM score in **Table 2** is the average of all the benchmarks used in the Huggingface open VLM leaderboard.

Table 2. Overview of VLMs tested, their components and score on popular benchmarks

| Name | Params | Vision Encoder | LLM | OVLM Score | MMBench | MMMU | MMVet |
|---|---|---|---|---|---|---|---|
| Aquila-VL2 | 2.2 B | SigLIP-So400M | Qwen 2.5 1.5B | 59.4 | 75.2 | 46.9 | 43.8 |
| Phi-3.5-vision | 3.8 B | CLIP ViT-L/14 | Phi 3.5 mini | 53 | 67.4 | 44.6 | 43.2 |
| Molmo 7B O 0924 | 7 B | CLIP ViT-L/14 | Olmo 7B | 54 | 62.9 | 42.9 | 50 |
| InternVL 2.5 | 8 B | InternViT-300M | InternVL 2.5 7B | 68.1 | 82.5 | 56.2 | 62.8 |
| LlavaOneVision | 8 B | SigLIP-So400M | Qwen 2 7B | 61.2 | 76.8 | 46.8 | 50.6 |
| Ovis 2 8B | 8 B | AIMv2-huge | Qwen 2.5 7B | | 83.3 | 59.0 | 68.5 |
| Qwen 2.5 VL | 8.3 B | own ViT | Qwen 2.5 7B | 70.4 | 82.6 | 56.2 | 66.6 |
| MiniCPM 2.6 o | 8.7 B | SigLIP-So400M | Qwen 2.5 7B | 70.2 | 80.6 | 50.9 | 67.2 |
| PaliGemma2 ft-docci | 10 B | SigLIP-So400M | Gemma 2 9B | - | - | - | - |
| Ovis1.6 Gemma2 | 10.2 B | SigLIP-So400M | Gemma 2 9B | 68.8 | 80.5 | 55 | 65 |
| Llama 3.2 Vision | 11 B | unclear | Llama 3.1 8B | 57.3 | 72.2 | 42.2 | 51.5 |
| Pixtral | 13 B | PixtralViT | Mistral Nemo | 61 | 72.7 | 51.1 | 58.5 |
| Ovis 2 16B | 16 B | AIMv2-huge | Qwen 2.5 14B | | 85.2 | 59.6 | 68.4 |
| Aria (MoE) | 25 B | SigLIP-So400M | no pretrained LLM | 64 | 77.4 | 54 | 56.9 |
| Ovis1.6 Gemma2 | 28 B | SigLIP-So400M | Gemma 2 27B | 71.3 | 82.2 | 60.3 | 68 |
| Ovis 2 34B | 34 B | AIMv2-1B | Qwen 2.5 32B | | 86.2 | 65.6 | 75.5 |
| InternVL 2.5 | 38 B | InternViT-6B | Qwen 2.5 32B | 73.5 | 85.4 | 64.6 | 67.2 |
| GPT-4o (1120) | | (as a reference) | | 71.6 | 80.5 | 69.9 | 75.1 |
| Claude 3.5 Sonnet | | (as a reference) | | 70.6 | 81.7 | 66.4 | 70.1 |

### 4.2 Jury as a Judge Evaluation

Evaluation of LLM or VLM answers is an issue as soon as you move beyond exact match and yes/no or multiple choice questions [14]. You can argue that adhering to output formatting instructions is equally important as semantic understanding and therefore it is rectified to score a model answer with zero points, despite answering semantically correct. However, we explicitly wanted to separate these two aspects and focus on the semantic correctness. Therefore, we decided to use the jury-as-a-judge approach [42], as an extension of and improvement over the popular LLM-as-a-judge evaluation [43]. We used the most capable models available that run on our own hardware and decided on Mistral 2 large 123B, Llama 3.3 70B and Qwen 2.5 72B as judges. They are very diverse (origin from the EU, the US and China) and all very capable with quality scores between 74 and 77 on the artificialananalysis.ai website (GPT-4o: 75). They all ran in a quantified version (AWQ) on a single H100 GPU. The VLMs were run with floating point 16-bit precision, except the biggest models, that were also run with AWQ quantization [44]. The VLMs were prompted once in German, asking for an answer in German and once in English, expecting an English answer. The temperature was set to 0.1 for all models, allowing for a bit of variation, but leading to relatively stable and therefore reproducible results. We analyzed the inter-rater agreement of the three judge models and found them to be high, with a Krippendorff's alpha of 0.992.

## 5 Results

The results in **Table 3** show that the capability of the models to identify the things depicted in the images varies a lot across categories. Whereas best models achieve over 90% on cars and products in both English and German language, performance drops to only 15% accuracy for German plants and 18% for German celebrities. English plants are identified in 48.8% in the best case, whereas English celebrities are identified in 88% from the best model, leading to the widest gap between English and German image contents. For animals the gap is wider between German and English prompting language (~20%) than between German and international image contents (~6-7%). Finally, sights are recognized worse in German language as well as for German image contents, but both differences are not equally pronounced as for animals, plants and celebrities. The best model overall is Lama 3.2 Vision 11B for English questions and answers and English image contents, whereas Qwen 2.5 VL 7B takes the lead when prompted in German language and asking for German image contents.

**Table 3.** Results of factual questions to VLMs

| VLM name | prompt lang | Animals | | Plants | | Celebrities | | Cars | | Products | | Sights | | Average | |
|---|---|---|---|---|---|---|---|---|---|---|---|---|---|---|---|
| | | de | en | de | en | de | en | de | en | de | en | de | en | de | en |
| Aquila-VL2 | de | 4.1% | 10.9% | 4.5% | 0.5% | 3.3% | 50.0% | 68.7% | 77.9% | 94.9% | 95.0% | 22.2% | 82.4% | 31.0% | 47.4% |
| | en | 18.2% | 23.2% | 4.9% | 10.8% | 2.6% | 49.5% | 67.9% | 79.0% | 94.9% | 94.1% | 27.8% | 88.3% | 37.2% | 56.5% |
| Aria 25B (MoE) | de | 32.9% | 41.1% | 14.0% | 18.6% | 9.2% | 78.3% | 88.4% | 80.8% | 92.3% | 94.6% | 64.4% | 94.7% | 49.7% | 62.8% |
| | en | 39.7% | 54.7% | 17.4% | 28.9% | 7.8% | 78.8% | 88.8% | 84.0% | 97.3% | **98.4%** | 58.9% | 95.7% | 51.8% | 69.7% |
| InternVL 2.5 38B | de | 18.0% | 25.7% | 7.2% | 7.7% | 5.2% | 49.5% | 73.2% | 63.0% | 92.3% | 96.7% | 54.4% | 86.7% | 41.9% | 51.5% |
| | en | 39.7% | 44.6% | 20.1% | 22.7% | 3.9% | 48.5% | 76.1% | 64.7% | 96.9% | 97.9% | 50.0% | 95.7% | 49.3% | 61.0% |
| InternVL 2.5 8B | de | 10.7% | 15.4% | 1.1% | 1.5% | 3.3% | 56.6% | 62.6% | 54.2% | 92.6% | 96.7% | 44.4% | 85.6% | 34.5% | 46.9% |
| | en | 25.4% | 31.6% | 9.8% | 12.4% | 2.4% | 58.6% | 63.7% | 54.6% | 94.0% | 96.7% | 42.2% | 92.0% | 39.8% | 55.3% |
| Llama-3.2 vision | de | 28.1% | 39.2% | 15.3% | 11.9% | 11.1% | 83.3% | 88.5% | 83.5% | 96.3% | 93.9% | 73.3% | 94.7% | 49.5% | 63.5% |
| | en | **59.3%** | 66.7% | 31.1% | 41.8% | 9.4% | 84.8% | 90.9% | 86.3% | 97.4% | 96.2% | 76.7% | 95.2% | 59.4% | **75.8%** |
| LlavaOneVision | de | 5.4% | 11.8% | 1.3% | 3.6% | 5.9% | 60.1% | 77.1% | **94.0%** | 88.6% | 94.8% | 37.8% | 84.6% | 35.7% | 53.2% |
| | en | 19.8% | 33.9% | 4.9% | 19.6% | 4.6% | 60.6% | 79.3% | **95.4%** | 91.5% | 95.8% | 33.3% | 94.7% | 41.4% | 62.7% |
| MiniCPM 2.6o | de | 10.5% | 15.2% | 2.7% | 6.2% | 3.3% | 26.3% | 73.7% | 69.5% | 90.1% | 90.3% | 47.8% | 87.2% | 36.9% | 44.9% |
| | en | 30.2% | 39.8% | 9.5% | 20.1% | 0.0% | 24.2% | 77.9% | 74.0% | 96.3% | 97.9% | 57.8% | 97.3% | 45.0% | 56.3% |
| Molmo-O-7B | de | 4.7% | 9.3% | 1.1% | 1.0% | 1.3% | 59.1% | 43.5% | 50.0% | 83.8% | 96.7% | 22.2% | 76.1% | 23.6% | 42.8% |
| | en | 11.0% | 23.4% | 4.0% | 11.3% | 2.6% | 61.6% | 39.9% | 47.8% | 86.4% | 93.4% | 17.8% | 74.5% | 29.2% | 50.2% |
| Ovis1.6-G2 9B | de | 37.7% | 43.0% | 14.0% | 20.1% | 11.1% | 64.6% | 82.5% | 80.6% | 95.5% | 97.2% | 50.0% | 93.1% | 50.2% | 63.7% |
| | en | 47.3% | 53.3% | 20.8% | 28.4% | 10.2% | 59.6% | 84.0% | 83.0% | **97.5%** | 98.4% | 50.0% | **97.3%** | 54.3% | 69.9% |
| Ovis1.6-G2 27B | de | 27.9% | 33.1% | 9.1% | 15.5% | 5.4% | 66.2% | 83.3% | 78.5% | 92.9% | 96.7% | 53.3% | 92.0% | 44.8% | 58.4% |
| | en | 36.6% | 47.6% | 14.4% | 22.7% | 4.1% | 63.6% | 84.1% | 79.9% | 96.6% | 97.9% | 49.2% | 95.2% | 49.2% | 65.5% |
| Paligemma2 | de | 15.9% | 20.7% | 9.8% | 18.0% | 5.4% | 38.4% | 86.4% | 76.1% | 90.1% | 91.3% | 33.3% | 81.4% | 40.6% | 51.9% |
| | en | 25.6% | 30.3% | 12.3% | 30.4% | 1.7% | 27.3% | 62.9% | 65.4% | 91.5% | 92.7% | 27.8% | 84.6% | 38.9% | 55.6% |
| Phi 3.5 vision | de | 9.5% | 15.8% | 1.5% | 2.1% | 4.6% | 62.1% | 67.3% | 54.4% | 88.6% | 90.6% | 24.4% | 74.5% | 32.6% | 45.5% |
| | en | 19.0% | 24.4% | 6.4% | 14.4% | 2.0% | 69.2% | 67.4% | 54.2% | 91.2% | 90.8% | 22.2% | 78.7% | 34.5% | 50.4% |
| Pixtral 12B | de | 20.6% | 29.7% | 5.3% | **22.7%** | 7.4% | 81.8% | **89.4%** | 83.0% | 94.0% | 95.8% | 65.6% | 84.6% | 44.1% | 58.3% |
| | en | 40.7% | 47.0% | 13.3% | 34.0% | 6.5% | 80.8% | **91.8%** | 84.5% | 96.6% | 97.0% | 67.8% | 90.4% | 49.9% | 66.1% |
| Qwen 2.5 VL 7B | de | 25.8% | 39.0% | **15.5%** | 21.6% | **18.3%** | **91.4%** | 89.8% | 84.5% | 95.7% | 96.7% | **81.1%** | **95.2%** | **53.5%** | 66.3% |
| | en | 56.8% | 56.6% | 28.4% | **45.4%** | **17.2%** | **92.9%** | 91.6% | 88.6% | 97.4% | 97.4% | **88.9%** | 96.8% | **61.0%** | 74.6% |

## 6 Discussion

On average, the models showed 13% worse performance on German image contents than on international ones. Therefore, our initial hypothesis is clearly supported. The difference between English prompts and answers and German prompts and answers was also significant, but smaller (3.9% for German images, 6.0% for the other). This means that English prompting slightly increases the gap between German image contents and international ones.

## 6.1 Results across Categories

The most significant differences in recognizing German **image contents** compared to international image contents appears for celebrities (56% on average) and sights (41%). For the other categories it was much less pronounced (20% for plants, 15% for animals). Answers in the products category can completely be given based on OCR. Results show that there is no significant difference in recognizing German text compared to English text and also the brand logos are recognized equally well (answers are composed of brand name and product name).

**Car names** are also language independent. Therefore, no significant difference could be found for German and English prompts. Image contents do make a difference for some models, but models that are better in recognizing German car contents and those better with English car contents are relatively balanced. Phi, InternVL, Aria, Pixtral, Qwen and Llama were better for German image contents, whereas LlavaOneVision, Aquila and Molmo were better for English image contents. The difference is only modest varying between 17% better for English to 13% better for German image contents. With an average of 77% and 74%, cars are generally recognized very well, only beaten by products (~94%) and international sights (~89%).

For **celebrities** the difference between recognizing German vs. international image contents becomes bigger, the better the model recognizes the international celebrities, because the German ones are recognized with only 18% accuracy at most, while the international ones are recognized in up to 93% of the cases, widening the gap to 75%. The secondary aspect researched was whether public datasets that are potentially included in the training data of models make a difference. Only PaliGemma2 shows signs supporting this hypothesis. German celebrities are recognized in 1.7% of cases, international ones from the self-collected dataset in 38% of cases and the public Celeb1k dataset in 63% of cases. For English prompts the gap between self-collected and public data becomes even 35% absolute, although the self-collected images are of course also publicly available and not proprietary to the paper authors. Smaller differences of 10% and 5% are visible for the MiniCPM and Qwen models. Other models show the opposite direction and recognize the self-collected images much better than the ones from Celeb1k (18%-28%).

For **sights**, the trend was the other way around compared to celebrities. The better models recognized the international sights, the less pronounced was the difference to German sights. Models with less than 83% accuracy on international sights showed 48% and more difference to the German images (54.2% on average). Models with 95% and more accuracy on international sights had an average of 31.4% difference to German sights. A notable exception is Pixtral, which shows a relatively small difference (~20%) while performing only mediocre compared to the other models. LavaOneVision and Aquila showed the highest difference with over 60% absolute.

## 6.2 Scientfic vs. English vs. German Names

In several cases, we found that the mapping of different representations to the same concept in latent space (Fig. 1) does not work effectively for the German language, as desired by the creators of the model. This is especially visible for animals in Qwen 2.5 7B, Llama 3.2 Vision 11B, Pixtral 12B and InternVL 2.5 38B. These models can correctly identify the animal in the picture in 50-70% of the cases according to its scientific name, but only in 18-28% of cases according to its German name (**Table 4**). This is a difference of 45% absolute for Qwen and still 32% for the other models, whereas the English name is identified nearly as well as the scientific one (0.2% to 14% difference).

**Table 4.** Comparison of accuracy on scientific and common animal names

| model name | prompt lang | scientific name | | common name | | Δ scientific - common | |
|---|---|---|---|---|---|---|---|
| | | de img | en img | de img | en img | de | en |
| InternVL 2.5 38B | de | 49.2% | 35.8% | 18.0% | 25.7% | 31.2% | 10.1% |
| | en | 43.8% | 34.9% | 39.7% | 44.6% | 4.1% | -9.7% |
| Llama 3.2 Vision 11B | de | 60.7% | 50.1% | 28.1% | 39.2% | 32.6% | 10.9% |
| | en | 59.1% | 50.3% | **59.3%** | **66.7%** | -0.2% | -16.4% |
| Ovis1.6 Gemma2 9B | de | 37.0% | 37.3% | **37.7%** | **43.0%** | -0.7% | -5.6% |
| | en | 37.0% | 35.8% | 47.3% | 53.3% | -10.3% | -17.5% |
| Ovis1.6 Gemma2 27B | de | 31.2% | 31.2% | 27.9% | 33.1% | 3.3% | -1.9% |
| | en | 30.6% | 32.0% | 36.6% | 47.6% | -6.0% | -15.6% |
| Paligemma2 | de | 1.0% | 2.7% | 15.9% | 20.7% | -14.9% | -18.0% |
| | en | 1.2% | 3.4% | 25.6% | 30.3% | -24.4% | -26.9% |
| Pixtral 12B | de | 52.5% | 47.8% | 20.6% | 29.7% | 31.9% | 18.1% |
| | en | 49.6% | 48.2% | 40.7% | 47.0% | 8.9% | 1.1% |
| Qwen 2.5 VL 7B | de | **71.3%** | 52.0% | 25.8% | 39.0% | 45.5% | 13.0% |
| | en | **70.9%** | **54.5%** | 56.8% | 56.6% | 14.1% | -2.1% |

The other models have significantly worse results already with identifying the animal correctly, no matter whether the scientific, English or German name should be named. Despite that, the German name is still recognized between 19% and 8% less than the scientific one. A notable exception are the Ovis 1.6 Gemma 2 models, which score very similar for scientific and German names (0.7-3.3% difference), which allowed the 9B model to take the lead for German animal names but is still roughly half as good as the best model for scientific names (71%). Another exception was Paligemma 2, which scored 16% for German names, 26% for English names, but only 1% for scientific names.

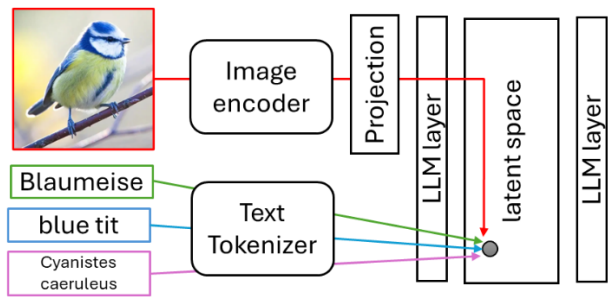

**Fig. 1.** Mapping of inputs to a common vector in latent space within a VLM (own illustration)

To dive deeper into this issue, we did a few tests with text only questions for the VLM (see **Table 5**). We asked the model to tell the corresponding German name for the given scientific name or for the given English name for some examples where the model correctly identified the scientific, but not the German name. No image input was provided for these queries. The column "image => German" is for reference and represents the model output for the normal question to identify the animal/plant on the image. As you can see, the model is not able to translate from the scientific name to the German name and only in one case can translate correctly from English to German. Astonishingly, the other way around from German to the scientific name is working in several cases. We verified this with the Llama 3.1 8B base LLM and got the same answers. The issue, therefore, appears to originate not from the VLM training but rather from the LLM pretraining.

Table 5. Examples for animals and plants correctly identified with scientific name

| Model | Ground-truth | Image => German | Scientific => German | English => German | German => Scientific |
|---|---|---|---|---|---|
| Llama | Buchfink | Goldammer ☹ | Trauerkehlchen ☹ | Gimpel ☹ | Fringilla coelebs ✓ |
| Llama | Stockente | Mallard ☹ (correct EN) | Pfeifente ☹ | Maulwasserente ☹ | Anas platyrhynchos ✓ |
| Llama | Bartnelke | Düfte Gartendieb ☹ | Widderdistel ☹ | Schwedenkönigskraut ☹ | Cornus sanguinea ☹ |
| Aria | Haubentaucher | Fischadler ☹ | Kolbenente ☹ | Gänsesäger ☹ | Not able to answer |
| Pixtral | Bachstelze | Weißkehlchen | Weißmeise ☹ | Bachstelze ✓ | Motacilla alba ✓ |

### 6.3 Do Bigger Models Perform Better?

The comparison between **differently sized LLM backends** reveals small gains of up to 14% at most and an average of 6% for InternVL (38B vs 8B). On the other hand, OvisGemma (27B vs 9B) showed stronger performance with the smaller model (3%). This controversial is hardened, as the overall performance of models on our dataset shows no clear relationship to the size of the model or the average OVLM score. Llama 3.2 Vision is one of the top performing models on our benchmark, but scores only 57.3 in the OVLM score, which makes it the third worst model there. Pixtral and Aria show similar results to Llama on our benchmark but score a bit better on public benchmarks.

**The biggest models**, Ovis 1.6 Gemma 2 27B and InternVL 2.5 38B are the best models on public benchmarks but show only mediocre performance on our dataset. MiniCPM 2.6o and InternVL 2.5 8B are among the worst models on our dataset, whereas they perform very well on OVLM. Molmo is the worst model on German images, and second worst on English images. The small models, Aquila and Phi 3.5 vision, are also among the worst models. The best overall model Qwen 2.5 is of medium size but also shows strong performance on public benchmarks (third place).

# 7    Limitations

Although the choice of models was guided by the criteria given, there remains a subjective aspect as we defined no strict rule that allows us to objectively decide about inclusion or exclusion for the VLMs to test.

Our focus on German image content and the desire for a balanced dataset led to the inclusion of some less well-known entities. This choice may limit the generalizability of our findings, as even German citizens might not be familiar with every German celebrity, animal, or sight in our dataset. Future work could involve establishing a human baseline with German participants to provide a more robust comparison.

While our dataset includes both uniquely German and internationally recognized entities, it is important to note that some contents can contain overlaps. For example, a few animals from the German subset also live in other European countries, or even in the US like the wolf. The same applies to cars. Although there are differences regarding cars typically found on German streets and those on US streets, German manufacturers sell their cars worldwide and some import cars like the Toyota RAV4 are popular both in the US as well as in Germany. However, our dataset still has some major differences, e.g. pickups like the Ford F-150 and brands like Buick, Chevrolet, Dodge and GMC that are rarely sold in Germany, but are typical for the US. The other way around, small cars like the Opel Corsa, Fiat 500, Renault Clio and Peugeot 208 are widespread in Germany and seldomly seen in the US, so that the datasets are still widely different.

Our benchmark prioritized usefulness for everyday users, leading us to score common names of animals and plants as correct rather than insisting on exact scientific species names. The decision about what is an acceptable generalization from the exact species and what is not, remained subjective and mainly rooted in experience with everyday conversations. Country-specific prefixes like "European" were seen as not necessary to count as correct. Therefore, "badger" for example, was scored as correct, although it is strictly speaking the "European badger" that we were looking for. Additionally, as a rule of thumb, you can take the number of species within one animal family. We insisted on more specific names for mammals and birds with few species in one animal family, whereas for insects with hundreds of species within the same family, we already considered the animal family name as correct, e.g., longhorn beetle or mayfly, because this is the one that is usually known by laymen. However, we were not satisfied with "duck", "butterfly" or "fish" as answers, as we considered those as too coarse. To improve rigor, future work could develop a more objective scoring rubric grounded in linguistic or cultural studies.

**Conclusion and Outlook**

In this work we presented an analysis of the accuracy of open-weight VLMs on factual knowledge in German and English language. We disentangle the image-related aspects from the textual ones by analyzing accuracy in both prompt languages and images from German or international contexts. We found that in the categories celebrities and sights, VLMs struggle because of lacking perceptual knowledge of the image contents, whereas in the categories animals and plants the tested models often are able to correctly identify the image contents according to the scientific name or English common name, but not according to the German name. We analyzed a few examples deeper and could trace back the mistakes to the LLM backbone, since neither the VLMs nor the underlying LLM alone was able to correctly state the German name given either the scientific or English counterpart.

For the categories cars and products from the supermarket, no significant differences in accuracy could be found regarding prompt language or image contents. For cars, this was attributed to the international car market whereas for products it seems to stem from the OCR-heavy nature of these images that allows models to answer the questions solely with OCR capabilities without additional need for visual understanding.

The analysis shows that although it is understandable to focus benchmarks on questions that require intricate understanding of both visual and textual parts or even require multi-step reasoning and extensive background knowledge, VLMs already struggle with simple factual knowledge as soon as image contents are a bit off mainstream. Therefore, we call for establishing benchmarking practices that also test image only questions as well as text only questions in addition to the typical VQA type of questions. This is an area that has been neglected up to now. Additionally, researchers should pay attention to include country-specific image and text contents into their benchmarks and compare performance across languages, since English language is still dominating for training of generative AI models and even high-resource languages like German are underrepresented in training data, which results in significant performance degradation for use of the models in the respective language.

## References

[1] J. Wu, W. Gan, Z. Chen, S. Wan, and P. S. Yu, "Multimodal Large Language Models: A Survey," In *2023 IEEE Int. Conf. on Big Data* (pp. 2247-2256). IEEE.

[2] J. Zhang, J. Huang, S. Jin, and S. Lu, "Vision-language models for vision tasks: A survey," *IEEE Trans. Pattern Anal. Mach. Intell.*, 2024

[3] J.-B. Alayrac *et al.*, "Flamingo: a visual language model for few-shot learning" *Advances in neural information processing systems, 35, 23716-23736.*.

[4] J. Li, D. Li, C. Xiong, and S. Hoi, "Blip: Bootstrapping language-image pre-training for unified vision-language understanding and generation," in *International Conference on Machine Learning*, in BLIP. PMLR, 2022, pp. 12888–12900.

[5] J. Li, D. Li, S. Savarese, and S. Hoi, "Blip-2: Bootstrapping language-image pre-training with frozen image encoders and large language models," *In International conference on machine learning (pp. 19730-19742). PMLR*.

[6] W. Dai *et al.*, "InstructBLIP: Towards General-purpose Vision-Language Models with Instruction Tuning," Jun. 15, 2023, *arXiv*: arXiv:2305.06500.

[7] Y. Qiao *et al.*, "Prism: A Framework for Decoupling and Assessing the Capabilities of VLMs," Jun. 20, 2024, *arXiv*: arXiv:2406.14544.

[8] "Open VLM Leaderboard - a Hugging Face Space by opencompass." Accessed: Feb. 13, 2025. [Online]. Available: https://huggingface.co/spaces/opencompass/open_vlm_leaderboard

[9] Z. Chen et al., "Expanding Performance Boundaries of Open-Source Multimodal Models with Model, Data, and Test-Time Scaling," Jan. 13, 2025, *arXiv*: arXiv:2412.05271.

[10] Q. Team, "Qwen2.5 VL! Qwen2.5 VL! Qwen2.5 VL!," Qwen. Accessed: Feb. 12, 2025. [Online]. Available: https://qwenlm.github.io/blog/qwen2.5-vl/

[11] A. I. Meta, "Llama 3.2: Revolutionizing edge AI and vision with open, customizable models," *Meta AI Blog Retrieved Dec.*, vol. 20, p. 2024, 2024.

[12] M. Deitke et al., "Molmo and PixMo: Open Weights and Open Data for State-of-the-Art Vision-Language Models," Dec. 05, 2024, *arXiv*: arXiv:2409.17146.

[13] P. Agrawal et al., "Pixtral 12B," Oct. 10, 2024, *arXiv*: arXiv:2410.07073.

[14] R. Peinl and J. Wirth, "Evaluation of Medium-Sized Language Models in German and English Language," *Int. J. Nat. Lang. Comput.*, vol. 13, no. 1, Feb. 2024.

[15] D. Romero et al., "CVQA: Culturally-diverse Multilingual Visual Question Answering Benchmark," Nov. 04, 2024, *arXiv*: arXiv:2406.05967.

[16] H. Wang et al., "M4U: Evaluating Multilingual Understanding and Reasoning for Large Multimodal Models," May 24, 2024, *arXiv*: arXiv:2405.15638.

[17] Z. Li, X. Wu, H. Du, H. Nghiem, and G. Shi, "Benchmark Evaluations, Applications, and Challenges of Large Vision Language Models: A Survey," Jan. 29, 2025, *arXiv*: arXiv:2501.02189.

[18] S. Antol et al., "Vqa: Visual question answering," in *Proceedings of the IEEE international conference on computer vision*, 2015, pp. 2425–2433.

[19] K. Marino, M. Rastegari, A. Farhadi, and R. Mottaghi, "Ok-vqa: A visual question answering benchmark requiring external knowledge," in *Proc. of the IEEE/cvf conference on computer vision and pattern recognition*, 2019, pp. 3195–3204.

[20] A. Singh et al., "Towards vqa models that can read," in *Proc. of the IEEE/CVF conference on computer vision and pattern recognition*, 2019, pp. 8317–8326.

[21] A. Masry, D. X. Long, J. Q. Tan, S. Joty, and E. Hoque, "ChartQA: A Benchmark for Question Answering about Charts with Visual and Logical Reasoning," Mar. 19, 2022, *arXiv*: arXiv:2203.10244.

[22] P. Lu et al., "MathVista: Evaluating Mathematical Reasoning of Foundation Models in Visual Contexts," 12th Int. Conf. on Learning Representations ICLR2024.

[23] W. Yu et al., "MM-Vet: Evaluating Large Multimodal Models for Integrated Capabilities," *41st Int. Conf. on Machine Learning*, PMLR 235:57730-57754, 2024.

[24] X. Yue et al., "Mmmu: A massive multi-discipline multimodal understanding and reasoning benchmark for expert agi," in *Proc. of the IEEE/CVF Conference on Computer Vision and Pattern Recognition*, 2024, pp. 9556–9567.

[25] Y. Wang et al., "MMLU-Pro: A More Robust and Challenging Multi-Task Language Understanding Benchmark," Nov. 06, 2024, *arXiv*: arXiv:2406.01574.

[26] T. Guan et al., "HallusionBench: an advanced diagnostic suite for entangled language hallucination and visual illusion in large vision-language models," in *Proc. of the IEEE/CVF Conference on Computer Vision and Pattern Recognition*, 2024, pp. 14375–14385.

[27] Y. Li, Y. Du, K. Zhou, J. Wang, W. X. Zhao, and J.-R. Wen, "Evaluating Object Hallucination in Large Vision-Language Models," In: *Proceedings of 2023 Conference on Empirical Methods in Natural Language Processing* (pp. 292-305).
[28] B. Li *et al.*, "SEED-Bench-2: Benchmarking Multimodal Large Language Models," Nov. 28, 2023, *arXiv*: arXiv:2311.17092.
[29] B. Li, Y. Ge, Y. Chen, Y. Ge, R. Zhang, and Y. Shan, "SEED-Bench-2-Plus: Benchmarking Multimodal Large Language Models with Text-Rich Visual Comprehension," Apr. 25, 2024, *arXiv*: arXiv:2404.16790.
[30] Y. Liu *et al.*, "MMBench: Is Your Multi-modal Model an All-Around Player?," in *Computer Vision – ECCV 2024*, vol. 15064, A. Leonardis, E. Ricci, S. Roth, O. Russakovsky, T. Sattler, and G. Varol, Eds., in Lecture Notes in Computer Science, vol. 15064. Cham: Springer Nature Switzerland, 2025, pp. 216–233.
[31] J. Tang *et al.*, "MTVQA: Benchmarking Multilingual Text-Centric Visual Question Answering," Nov. 19, 2024, *arXiv*: arXiv:2405.11985.
[32] R. J. Das, S. E. Hristov, H. Li, D. I. Dimitrov, I. Koychev, and P. Nakov, "EXAMS-V: A Multi-Discipline Multilingual Multimodal Exam Benchmark for Evaluating Vision Language Models," Mar. 15, 2024, *arXiv*: arXiv:2403.10378.
[33] H.-L. Sun *et al.*, "Parrot: Multilingual Visual Instruction Tuning," Aug. 11, 2024, *arXiv*: arXiv:2406.02539.
[34] L. Li *et al.*, "M$^3$IT: A Large-Scale Dataset towards Multi-Modal Multilingual Instruction Tuning," *ArXiv Prepr. ArXiv230604387*, 2023.
[35] J. Pfeiffer *et al.*, "xGQA: Cross-Lingual Visual Question Answering," Mar. 17, 2022, *arXiv*: arXiv:2109.06082.
[36] C. Park *et al.*, "Translation Deserves Better: Analyzing Translation Artifacts in Cross-lingual Visual Question Answering," Jun. 04, 2024, *arXiv*: arXiv:2406.02331.
[37] S. Nayak *et al.*, "Benchmarking Vision Language Models for Cultural Understanding," Oct. 14, 2024, *arXiv*: arXiv:2407.10920.
[38] G. I. Winata *et al.*, "WorldCuisines: A Massive-Scale Benchmark for Multilingual and Multicultural Visual Question Answering on Global Cuisines," Feb. 08, 2025, *arXiv*: arXiv:2410.12705.
[39] A. Vayani *et al.*, "All Languages Matter: Evaluating LMMs on Culturally Diverse 100 Languages," Nov. 26, 2024, *arXiv*: arXiv:2411.16508.
[40] N. Du *et al.*, "Glam: Efficient scaling of language models with mixture-of-experts," in *Int. Conf. on Machine Learning (pp. 5547-5569)* 2022.
[41] D. Li *et al.*, "Aria: An Open Multimodal Native Mixture-of-Experts Model," Jan. 10, 2025, *arXiv*: arXiv:2410.05993.
[42] P. Verga *et al.*, "Replacing Judges with Juries: Evaluating LLM Generations with a Panel of Diverse Models," May 01, 2024, *arXiv*: arXiv:2404.18796.
[43] H. Huang *et al.*, "An Empirical Study of LLM-as-a-Judge for LLM Evaluation: Fine-tuned Judge Model is not a General Substitute for GPT-4," Nov. 05, 2024, *arXiv*: arXiv:2403.02839.
[44] J. Lin, J. Tang, H. Tang, S. Yang, X. Dang, and S. Han, "AWQ: Activation-aware Weight Quantization for LLM Compression and Acceleration," *Proceedings of Machine Learning and Systems*, *6*, 87-100. 2024